\documentclass[a4paper,10pt]{article}
\usepackage[utf8]{inputenc}
\usepackage{microtype,amsmath,amssymb}
\usepackage[left=31mm,right=30mm,top=22mm,bottom=19mm,footskip=10mm]{geometry}
\usepackage{xcolor,graphicx,booktabs,tabularx,array,caption,float}
\usepackage{titlesec,fancyhdr,tikz}
\newfloat{algorithm}{tbp}{loa}
\floatname{algorithm}{Algorithm}
\usetikzlibrary{arrows.meta,positioning,shapes.geometric}
\definecolor{navy}{HTML}{173F5F}
\definecolor{ruleblue}{HTML}{AEC5D7}
\definecolor{scanblue}{HTML}{E7EFF5}
\definecolor{routeblue}{HTML}{E9F1F2}
\definecolor{movegreen}{HTML}{E7F1EA}
\definecolor{waitgold}{HTML}{F7F0E2}
\definecolor{stopgray}{HTML}{F0F1F2}
\usepackage[colorlinks=true,linkcolor=navy,citecolor=navy,urlcolor=navy]{hyperref}
\hypersetup{pdftitle={Anytime-Valid Sequential Verification for Sensing-Efficient Robot Navigation under Perception Uncertainty},pdfauthor={Saad Korchi; Abderrahim Waga}}
\titleformat{\section}{\large\bfseries\color{navy}}{\thesection.}{.5em}{}
\titleformat{\subsection}{\normalsize\bfseries}{\thesubsection}{.5em}{}
\titlespacing*{\section}{0pt}{16pt}{8pt}
\titlespacing*{\subsection}{0pt}{11pt}{6pt}
\begin{document}
\begin{center}
{\fontsize{16}{20}\selectfont\bfseries\color{navy}
Anytime-Valid Sequential Verification for Sensing-Efficient\\
Robot Navigation under Perception Uncertainty\par}
\vspace{12pt}
{\large\bfseries Saad Korchi\textsuperscript{*}, Abderrahim Waga\par}
\vspace{8pt}
{\small Euromed University of Fez, Fez, Morocco\par}
\vspace{7pt}
{\small\color{navy}\texttt{saad.korchi@eidia.ueuromed.org}\hspace{18pt}\texttt{a.waga@ueuromed.org}\par}
\end{center}
\vspace{17pt}{\color{ruleblue}\hrule}\vspace{15pt}
\section*{Abstract}
When a robot navigates an unfamiliar grid, noisy sensor readings can make it hard to tell whether the cells along its next move are free. Additional scans can reduce that uncertainty, but a fixed scan count spends the same sensing effort at every position. In a static-grid simulation, a sequential rule uses a Gaussian likelihood ratio to determine when the cells newly covered by the robot's footprint have enough evidence of free space. On 324 held-out scenarios from 18 public maps, the sequential gate and a four-scan reference chosen on validation data both completed every mission without an observed collision. With their respective validation-selected routing weights, the gate averaged 123.15 scans per episode versus 224.13 for the reference, a reduction of 45.05\% (95\% map-stratified bootstrap interval: 42.21--48.95\%). In a retrospective check on the same scenarios with routing weights held equal, the reductions were 43.91\% and 43.37\% at the two tested weights, with all missions completed. On 216 scenarios from 12 previously unused maps, allocating more of the same certification error budget near the initial start--goal rectangle cut the sequential gate's scan count by a further 11.58\%. The spatial variant completed 206 missions versus 208 for the original gate and increased computation time. The collision bound depends on the assumed observation model and footprint geometry. In a stress test with persistent per-cell bias, the original gate completed 0 of 54 missions.

\noindent\textbf{Keywords:} Active Sensing; Autonomous Mobile Robots; Grid-Based Path Planning; Occupancy Grid Mapping; Probabilistic Safety; Safe Navigation.

\section{Introduction}
Route planning and move verification answer different questions. A planner can find a short route through cells that have barely been observed, while taking the same number of scans at each position can delay progress along a well-observed route. The following experiments ask how many scans a robot can save by verifying only the cells that its next move would newly occupy.

At each position, the robot updates evidence for visible cells, plans with a shared occupancy-dependent route cost, and tests the next swept footprint. If any newly entered cell lacks a certificate, it observes again or stops at a fixed scan limit. The policies share routing code and recovery logic, but their validation-selected routing weights differ. The comparison therefore includes matched-weight controls.

The evaluation compares this rule with four fixed scan counts, a posterior threshold, and routing controls on held-out MovingAI maps. A retrospective comparison at matched routing weights helps distinguish the effect of sequential sensing from the effect of route selection. A separate experiment on a different set of maps tests whether concentrating a fixed certification error budget near the initial start--goal rectangle reduces scan counts. The contribution lies in how the navigation policy uses established sequential tests and union bounds, and in the measured tradeoff between sensing and mission completion. Results are reported separately for the two test sets, which contain different mixes of map families.

The specific contributions are:
\begin{enumerate}
\setlength{\itemsep}{1pt}
\item a move-level execution gate with an anytime-valid, cell-wise free-space certificate and an explicit conditional bound on episodic collision probability;
\item a paired benchmark comparing adaptive verification with fixed repeated sensing, including a retrospective matched-routing diagnostic to assess the role of routing weight;
\item a fixed spatial redistribution of the same certification budget, evaluated on previously unused maps with sensing, completion, and computation reported together; and
\item explicit stress tests and validation-set sensitivity analyses that identify where the model-based guarantee and practical completion diverge.
\end{enumerate}

The remainder of the paper is organized as follows. Section~2 places the method in the context of grid navigation, active sensing, and sequential inference. Section~3 describes the benchmark and observation model; Sections~4 and 5 specify the policies and their execution rule. Section~6 presents the benchmark, spatial-allocation, and robustness results. Section~7 examines parameter sensitivity, followed by discussion and limitations in Section~8 and reproducibility details in Section~9.

\section{Related Work}
\subsection{Grid-based planning and probabilistic environment representations}
A* supplies the graph-search foundation \cite{hart}, while occupancy grids make the spatial evidence for and against free space explicit \cite{elfes}. The planner uses an occupancy-dependent route cost to choose where to travel; a separate gate decides when the next movement may execute. This distinction matters because a route with low estimated risk does not by itself certify the cells swept by the robot.

\subsection{Navigation under obstacle uncertainty}
Safety under uncertain geometry has been studied beyond occupancy-grid routing. Axelrod, Kaelbling, and Lozano-P\'erez develop probabilistic safety reasoning for trajectories and policies under imperfect obstacle observations, including a safe variant of RRT \cite{axelrod}. Their work illustrates why uncertainty assumptions and the meaning of a safety statement must be explicit. The experiments here focus on static grids with exact localization and cardinal motion.

Active perception studies jointly consider where to move and what to observe. Popovi\'c et al. survey adaptive informative path planning under resource limits \cite{popovic}. Nguyen et al. couple predicted collision and information gain in visually attentive navigation, with simulation and physical flight experiments \cite{nguyen}. Those settings optimize viewpoints and motion under richer sensing and dynamics. The intervention studied here leaves the sensor's viewing geometry fixed and adaptively repeats measurements before a discrete move. Fixed-scan controls provide a reference for measuring that intervention within the simulator. Stephens et al. study safe exploration when environmental hazards, and in one setting the map itself, must be learned; they combine Gaussian-process predictions with decision-process planning and report a physical-robot trial \cite{stephens}. Dawson et al. optimize chance-constrained trajectories under uncertain obstacle geometry and robot tracking error \cite{dawson}. Both address safety under uncertainty, but their exploration objectives, motion models, and risk formulations differ from repeated cell measurements before a discrete move. Neither is an empirical baseline here; the comparisons below isolate the execution gate within a common grid simulator.

\subsection{Sequential evidence and time-uniform validity}
When the number of observations is chosen adaptively, a decision rule must account for repeated inspection of accumulating evidence. Howard et al.\ develop time-uniform inference using nonnegative process techniques \cite{howard}. Here, a simple likelihood-ratio martingale supplies a cell-wise boundary that can be checked after every scan. A fixed allocation of error across all map cells then yields a conditional episodic bound. The bound applies established sequential reasoning to the simulator's observation model.

\subsection{Benchmarking and experimental interpretation}
Public pathfinding maps and official query files make geometry and scenario provenance inspectable. The MovingAI benchmark collection supports consistent comparisons across grid-based planning studies \cite{sturtevant}. The experimental design specifies the sensor model, robot footprint, and closed-loop execution policy separately from the maps. The evaluation retains original map geometry and records query eligibility, source hashes, seeds, and split membership. The benchmark remains conditional on a selected set of reachable, moderate-length missions.

\section{Dataset and Experimental Design}
\subsection{Dataset composition}
The initial study uses 30 full MovingAI parents: ten room maps, ten maze maps, and ten Dragon Age Origins (DAO) maps. Two parents per family calibrate the routing posterior, two select parameters, and six form the held-out test. Each evaluation parent contributes six official queries at three noise levels. The follow-up reuses the existing validation set and adds 12 previously unused test parents: six maze512-4 maps and six DAO maps.

\begin{table}[H]
\caption{Experimental partitions. Counts refer to navigation scenarios; calibration uses separate visible-cell observations on six parents.}
\centering\small
\begin{tabular}{lrrl}
\toprule
Partition & Parents & Scenarios & Purpose\\
\midrule
Initial calibration & 6 & --- & Fit routing posterior\\
Initial validation & 6 & 108 & Select policy settings\\
Initial held-out test & 18 & 324 & Eight-policy comparison\\
Follow-up validation & Same 6 & Same 108 & Select spatial width\\
Follow-up held-out test & 12 new & 216 & Nine-policy comparison\\
\bottomrule
\end{tabular}
\end{table}

\subsection{Scenario preparation and observation model}
Two official queries are selected from each published octile-distance band $[12,32)$, $[32,52)$, and $[52,72)$. Eligibility requires footprint-valid endpoints, four-connected reachability, and a recomputed four-connected optimum of at most 180 steps. The official diagonal distance is used only for stratification. Maps are not cropped, rescaled, or closed with artificial walls. Room and initial maze maps have $512\times512$ cells; DAO dimensions vary.

The room maps considered for the follow-up could not meet the query quotas once the robot's footprint was taken into account. Both failed preparation attempts and the final two-family design were recorded before any follow-up policy outcomes were available. The resulting change in map families limits comparisons across experiments.

For the 12 accepted follow-up maps, the official scenario files contain 900 queries per family in the three distance bands. The deterministic search examines candidates only until it fills each two-query band quota: 46 DAO candidates (36 selected, seven footprint rejections, three connectivity rejections) and 143 maze candidates (36 selected, 106 footprint rejections, one connectivity rejection). These are counts of candidates inspected before quota completion, not pass rates for all 1,800 available band-matched queries. Both attempted room-map designs failed the same quotas before policy evaluation.

The robot is a disk of radius 0.3\,m on a grid with 0.5\,m cells. It moves one cardinal step at a time with exact pose and actuation. The simulator casts rays out to five cells (2.5\,m) and stops each ray at the first occupied cell on the ground-truth map. The policies receive the visible cell indices and noisy readings. The simulator deduplicates cells within each scan, and each visible cell supplies
\begin{equation}
Z_{it}\mid O_i \sim \mathcal{N}(2O_i-1,\sigma^2),\qquad O_i\in\{0,1\},
\end{equation}
under the nominal conditional-independence model. The standard deviation is known, with $\sigma\in\{0.6,0.9,1.2\}$. A deterministic noise field indexed by episode seed, scan index, and absolute cell supplies common random conditions; different policies may visit different cells at different times.

\section{Navigation Policies and Parameter Selection}
\subsection{Shared routing and fixed repeated sensing}
Evidence accumulates as $E_i=\sum_t 2Z_{it}/\sigma^2$. A logistic map $p_i=\operatorname{sigmoid}(aE_i+b)$, fitted only on calibration parents, provides occupancy estimates for routing. An inflated hard threshold of 0.65 excludes observed high-occupancy cells. A footprint-based score $q_v$ sums marginal occupancy, with clipping, and weighted A* uses an edge cost of the form $1+\lambda[-\log(1-q_v)]$.

Uniform-$k$ collects $k\in\{1,2,4,8\}$ scans at each position before movement. The principal policies share four-step route commitment and observe after each move. Their selected routing weights differ: $\lambda=1$ for Anytime-gate and $\lambda=4$ for Uniform-4. Shortest A* and risk-tie A* are descriptive controls. Expanding computational windows limit planning cost, with a full-map fallback on infeasibility; a feasible windowed route need not minimize the global weighted objective.

\subsection{Sequential and posterior execution gates}
Anytime-gate starts with one scan and waits until every newly entered footprint cell has received a free-space certificate. Its original allocation is uniform across the full map. Spatial-gate changes only the fixed cell-wise allocation, placing more of the error allowance near the rectangle defined by the initial start and goal. Posterior-gate instead waits until the summed posterior occupancy of new footprint cells is at most a tuned threshold; it is not assigned the sequential gate's episodic bound.

\subsection{Validation and frozen configurations}
The initial validation evaluates 24 configurations: 12 uniform policies, nine posterior-gate settings, and three routing weights for Anytime-gate. Selection prioritizes success, then fewer collisions and scans. Uniform-4 ($\lambda=4$) is selected as the primary fixed-sensing reference; Anytime-gate selects $\lambda=1$. The follow-up fixes these settings and tests spatial widths $h\in\{8,16,32\}$ on the same validation set; all complete 108/108 missions, and $h=8$ uses the fewest scans. Source, configuration, and manifests are hashed before final testing. These hashes document the internal configuration freeze.

\begin{table}[H]
\caption{Shared execution settings and the principal verification policies.}
\centering\small
\begin{tabularx}{\linewidth}{@{}lX@{}}
\toprule
Setting & Value or rule\\
\midrule
Robot / grid & Radius 0.3\,m; resolution 0.5\,m; cardinal translations\\
Observation & Range five cells; occlusion; known Gaussian noise scale\\
Execution limits & 4,096 scans; 512 moves; 32 stationary scans\\
Routing commitment & Up to four steps, invalidated by blocked next footprint\\
Uniform-4 & Four scans at each position; routing weight $\lambda=4$\\
Anytime-gate & Uniform error allocation; $\delta=0.05$; $\lambda=1$\\
Spatial-gate & Fixed spatial allocation; $\delta=0.05$; $h=8$; $\lambda=1$\\
\bottomrule
\end{tabularx}
\end{table}

\section{Execution Verification and Simulation Workflow}
\subsection{Cell certificates and conditional safety}
For an occupied cell, let $M_i(n)=\exp[-E_i(n)]$. Under the stated conditional Gaussian model,
\begin{equation}
\mathbb{E}\!\left[\exp(-2Z_{it}/\sigma^2)\mid\text{past},O_i=1\right]=1.
\end{equation}
Indeed, the Gaussian moment-generating function gives $\exp(-2/\sigma^2+(2/\sigma^2)^2\sigma^2/2)=1$. Thus $M_i$ is a nonnegative martingale for an occupied cell under conditionally independent observations, even if the times at which that cell is observed depend on past data. For a fixed error allowance $\alpha_i$, Ville's inequality bounds the probability of ever crossing $1/\alpha_i$ by $\alpha_i$. Certification is implemented as
\begin{equation}
E_i(n)\leq-\log(1/\alpha_i),\qquad \alpha_i=\delta w_i,\quad \sum_i w_i=1.
\end{equation}
Anytime-gate sets $w_i=1/N$, where $N$ includes every map cell. Spatial-gate uses
\begin{equation}
w_i=\frac{0.1}{N}+0.9\frac{g_i}{\sum_jg_j},
\qquad g_i=\exp\!\left(-\frac{d_i^2}{2h^2}\right),
\end{equation}
where $d_i$ is distance to the initial start--goal bounding rectangle. Weights depend only on map dimensions and endpoints. They are set before the first observation and remain fixed throughout the episode.

The bound applies when observations are selected before the current scan's Gaussian draws. Visibility depends on the fixed map and a pose determined by prior observations. Ground truth determines the ray stops and the Gaussian means in the simulator; policy decisions use the resulting indices and readings.

For the swept footprint, measure distances in grid cells and place the centers of a horizontal unit step at $(0,0)$ and $(1,0)$. An obstacle cell is a unit square centered at integer coordinates $(x,y)$. Its vertical distance to the motion segment is $\max(|y|-1/2,0)$. If $x$ is 0 or 1, the square contains the corresponding endpoint's horizontal coordinate. If $x\leq-1$ or $x\geq2$, the closest horizontal point of the segment is the nearer endpoint. In both cases, the distance from the square to the segment equals the smaller of its distances to the two endpoint centers. The union of the two endpoint footprints therefore gives the swept cells for the disk radius of 0.6 cells. The vertical case follows by exchanging the axes. The planner excludes footprints that cross the map boundary.

The gate starts with a truly free initial footprint. It requires a certificate for every cell in the next footprint that the robot has not previously covered. Previously covered cells remain marked as free because occupancy is static. Before a first collision, all such cells are free, so a collision during an admitted move requires a false certificate for an occupied cell. A union bound gives $\Pr(\text{collision in one simulated episode})\leq\Pr(\text{any occupied cell certified free})\leq\sum_{i:O_i=1}\alpha_i\leq\delta$ under the stated model.

\subsection{Closed-loop implementation and outcome recording}
Each cycle gathers observations, updates evidence, checks the committed route, and evaluates the next footprint. An unresolved gate requests another scan or stops at the stationary limit; there is no unverified fallback. Planning failures receive the same recovery opportunity across policies. Success, collision, planning failure, verification stop, and timeout are mutually exclusive.

\begin{figure}[H]
\centering
\begin{tikzpicture}[>=Stealth, line width=.65pt, font=\small,
  flowbox/.style={draw=navy!70,rounded corners=2pt,align=center,minimum height=9mm,text width=5.2cm},
  branch/.style={draw=navy!75,diamond,aspect=3.3,align=center,inner sep=1pt,text width=3.7cm,fill=scanblue},
  outcome/.style={draw=navy!70,rounded corners=2pt,align=center,minimum height=11mm,text width=4.0cm}]
\node[flowbox,fill=scanblue] (scan) at (0,0) {Acquire one local scan};
\node[flowbox,fill=scanblue] (evidence) at (0,-1.05) {Update cell evidence and certificates};
\node[flowbox,fill=routeblue] (route) at (0,-2.10) {Check route and next footprint};
\node[branch] (gate) at (0,-3.55) {\mbox{All new footprint cells}\\\mbox{certified?}};
\node[outcome,fill=movegreen] (move) at (-3.0,-5.05) {Move to the next cell\\If needed, scan again};
\node[outcome,fill=waitgold] (wait) at (3.0,-5.05) {Remain in place\\Scan again if allowed};
\node[outcome,fill=stopgray,minimum height=8mm] (stop) at (3.0,-6.35) {At 32 scans: stop verification};
\draw[->] (scan)--(evidence);
\draw[->] (evidence)--(route);
\draw[->] (route)--(gate);
\draw[->] (gate.west)-| (move.north);
\draw[->] (gate.east)-| (wait.north);
\draw[->] (move.west)--(-5.45,-5.05)--(-5.45,.65)--(scan.north west);
\draw[->] (wait.east)--(5.45,-5.05)--(5.45,.65)--(scan.north east);
\draw[->] (wait)--(stop);
\end{tikzpicture}
\caption{Closed-loop sensor--action workflow for the sequential gates. A successful move returns to sensing unless the goal has been reached; the global move and scan limits, planning failures, and collisions also end an episode. Ground truth is used only by the observation generator and collision evaluator.}
\end{figure}
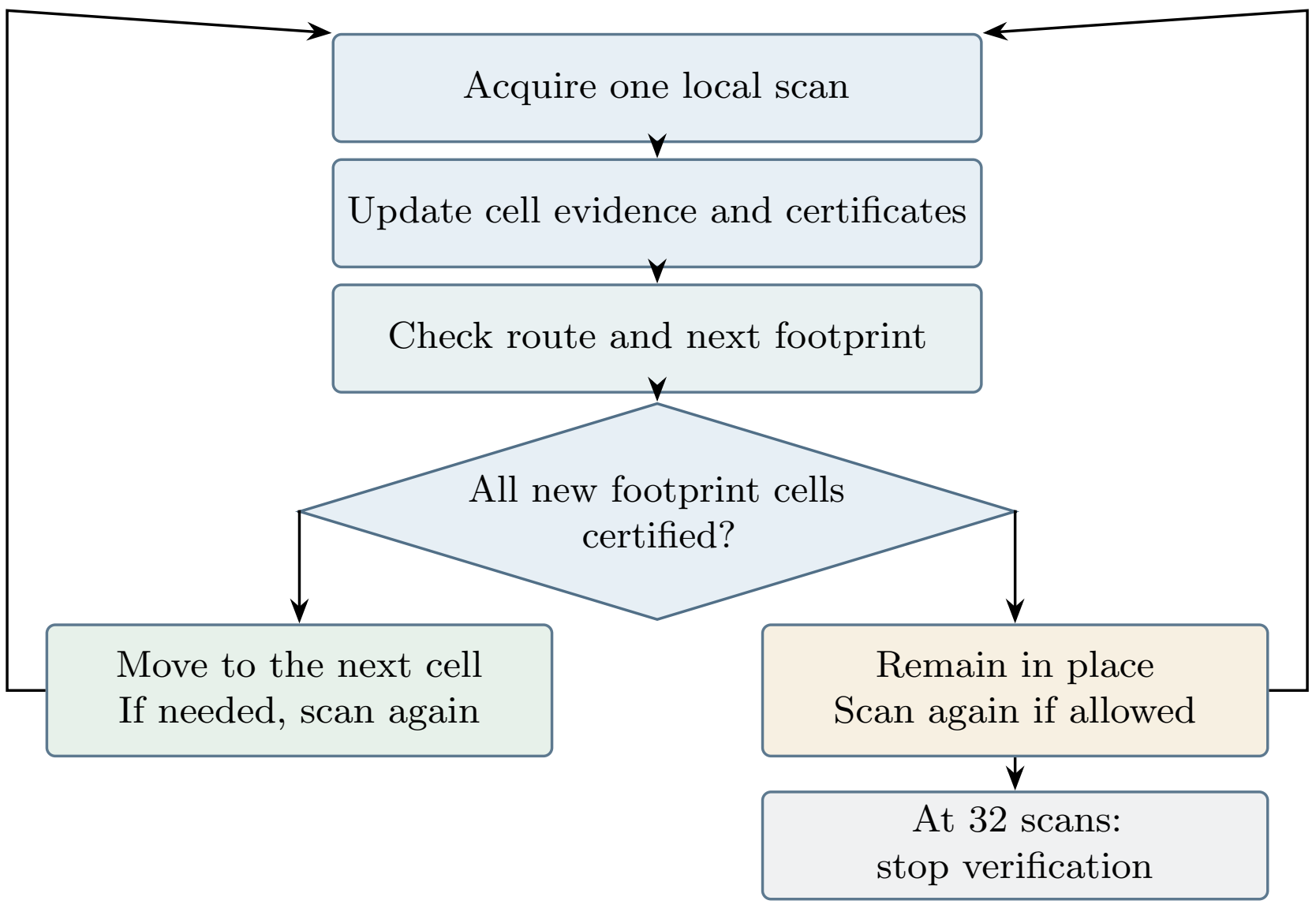

\begin{algorithm}[H]
\caption{Shared execution loop for Anytime-gate and Spatial-gate. The two policies differ only in the preassigned cell weights $w_i$. Planning and sensing use observations; collision checks use ground truth only in the evaluator.}
\label{alg:execution}\centering\small
\renewcommand{\arraystretch}{1.22}
\begin{tabularx}{\linewidth}{|>{\centering\arraybackslash}p{.72cm}|X|}
\hline
\multicolumn{2}{|p{\dimexpr\linewidth-2\tabcolsep-2\arrayrulewidth\relax}|}{\textbf{Inputs:} map dimensions, start and goal, fixed $\delta$, noise scale $\sigma$, routing weight $\lambda$, gate type.}\\ \hline
1 & Before sensing, fix $w_i=1/N$ (Anytime) or the start--goal spatial weights (Spatial); set the certification condition $E_i\leq-\log[1/(\delta w_i)]$. Initialize evidence $E_i=0$, certificates, and the known-free initial footprint.\\ \hline
2 & At the current pose, acquire one occlusion-limited scan. For each visible cell, add $2Z_{it}/\sigma^2$ to $E_i$ and permanently certify it free if its evidence crosses the threshold.\\ \hline
3 & Recheck the committed route; if needed, plan using the shared occupancy-dependent cost, retaining up to four forthcoming steps. If no feasible route emerges within 32 stationary scans, record planning failure.\\ \hline
4 & For the next step, find footprint cells not previously covered. If every such cell is certified, execute the step, mark its footprint known, and return to step 2.\\ \hline
5 & Otherwise, repeat from step 2 while fewer than 32 scans have been taken at this pose; then record verification stop. End at the goal, on collision, or at the global scan/move cap.\\ \hline
\end{tabularx}
\end{algorithm}

\section{Experimental Findings}
\subsection{Principal benchmark}
All eight methods run on the same 324 held-out scenarios (Table~\ref{tab:initial}). Uniform-4 is the primary reference selected on validation data. Anytime-gate, Uniform-4, and Uniform-8 each finish all 324 missions without an observed collision. Anytime-gate averages 123.15 scans per episode, compared with 224.13 for Uniform-4. The reduction is 45.05\% (95\% map-stratified bootstrap interval: 42.21--48.95\%). The paired mean difference is $-100.98$ scans, with a 95\% interval of $[-114.88,-90.04]$.

\begin{table}[H]
\caption{Initial held-out results: 324 scenarios from 18 parent maps. Length is averaged over successful episodes only.}
\label{tab:initial}\centering\small
\begin{tabular}{lrrrr}
\toprule
Method & Success (\%) & Collision (\%) & Scans & Length (m)\\
\midrule
A* & 88.27 & 11.73 & 48.48 & 25.65\\
Risk-tie A* & 90.74 & 9.26 & 50.16 & 25.80\\
Uniform-1 & 91.36 & 8.64 & 50.52 & 25.89\\
Uniform-2 & 98.77 & 1.23 & 107.94 & 27.10\\
Uniform-4 & 100.00 & 0.00 & 224.13 & 28.02\\
Uniform-8 & 100.00 & 0.00 & 433.28 & 27.08\\
Posterior-gate & 97.22 & 2.78 & 55.34 & 26.62\\
Anytime-gate & 100.00 & 0.00 & 123.15 & 27.50\\
\bottomrule
\end{tabular}
\end{table}

Because both principal policies finish every scenario, early termination cannot account for their scan difference. Cell-level measurements also fall by 44.82\% (95\% interval: 42.52--47.82\%). Mean successful path length changes from 28.02\,m under Uniform-4 to 27.50\,m under Anytime-gate. Uniform-2 uses only 107.94 scans on average, but fails four missions. This comparison shows why acquisitions and completion must be read together.

\begin{figure}[tbp]
\centering\includegraphics[width=\linewidth]{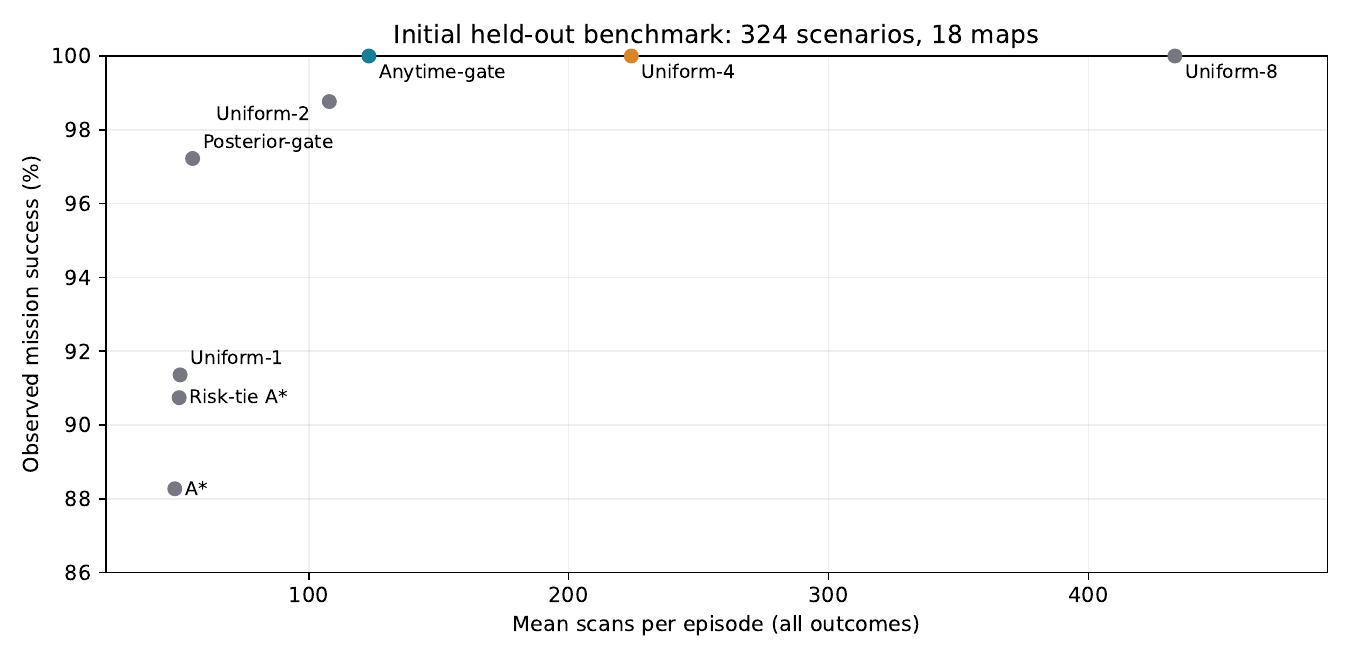}
\caption{Initial acquisition--success comparison. Each point summarizes all 324 scenarios. The truncated vertical range shows differences in observed completion; the plot does not establish equal underlying success probabilities.}
\end{figure}

The analysis pairs policies by scenario and uses 5,000 bootstrap replicates that resample whole maps within each family. Anytime-gate used fewer scans on average than Uniform-4 on each of the 18 test maps. When no event is observed, the bootstrap interval can collapse to a point, leaving equivalence, noninferiority, and population risk unresolved. The archived statistical outputs also include the original Holm-adjusted Monte Carlo sign-flip tests for success, collisions, and scans.

\begin{table}[H]
\caption{Post-hoc matched-routing diagnostic on the original 324 test scenarios. All four policies completed 324 missions with no observed collisions. The intervals resample 18 map parents within family; these analyses were specified after the original test outcomes were examined.}
\label{tab:matched}\centering\small
\begin{tabular}{lrrrr}
\toprule
Routing weight & Gate scans & Four-scan scans & Reduction (\%) & 95\% interval (\%)\\
\midrule
$\lambda=1$ & 123.15 & 219.56 & 43.91 & 42.57--45.16\\
$\lambda=4$ & 126.93 & 224.13 & 43.37 & 42.12--44.68\\
\bottomrule
\end{tabular}
\end{table}
Exploratory stratification of the matched $\lambda=1$ comparison shows reductions of 45.46\%, 43.52\%, and 42.78\% in DAO, maze, and room maps, respectively. By sensor noise ($\sigma=0.6,0.9,1.2$), they are 70.84\%, 47.46\%, and 13.58\%. Each stratum contains 108 scenarios, and both methods finish every one. The descriptive breakdown shows that the scan savings vary considerably with sensor noise.

\begin{table}[H]
\caption{Exploratory noise strata for the retrospective matched $\lambda=1$ comparison. Each row contains 108 scenarios; both policies finish 108. Scans include all outcomes.}
\centering\small
\begin{tabular}{lrrr}
\toprule
$\sigma$ & Anytime scans & Uniform-4 scans & Reduction (\%)\\
\midrule
0.6 & 64.48 & 221.11 & 70.84\\
0.9 & 113.50 & 216.03 & 47.46\\
1.2 & 191.47 & 221.56 & 13.58\\
\bottomrule
\end{tabular}
\end{table}

\subsection{Spatial allocation}
The second test uses 12 previously unused parents and 216 scenarios. The eight existing policies retain their original settings; only the spatial width is selected using validation data. Spatial-gate finishes 206 missions and times out on ten, without an observed collision. Anytime-gate finishes 208, with eight timeouts and no collisions; Uniform-4 also finishes 208, with six timeouts and two collisions (Table~\ref{tab:followup}).

\begin{table}[H]
\caption{Follow-up outcomes on 12 new parents. Other denotes planning failures; no verification stops occurred. Acquisitions include every outcome.}
\label{tab:followup}\centering\small
\begin{tabular}{lrrrrr}
\toprule
Method & Arrivals & Collisions & Timeouts & Other & Scans\\
\midrule
A* & 128 & 82 & 5 & 1 & 55.54\\
Risk-tie A* & 154 & 58 & 4 & 0 & 62.91\\
Uniform-1 & 154 & 58 & 4 & 0 & 61.93\\
Uniform-2 & 194 & 16 & 6 & 0 & 155.58\\
Uniform-4 & 208 & 2 & 6 & 0 & 332.65\\
Uniform-8 & 207 & 0 & 9 & 0 & 712.19\\
Posterior-gate & 182 & 27 & 7 & 0 & 81.07\\
Anytime-gate & 208 & 0 & 8 & 0 & 188.45\\
Spatial-gate & 206 & 0 & 10 & 0 & 166.63\\
\bottomrule
\end{tabular}
\end{table}

Spatial-gate reduces acquisitions by 11.58\% relative to Anytime-gate (95\% interval: 7.64--14.98\%) and by 49.91\% relative to Uniform-4 (41.12--55.88\%). The completion difference against Anytime-gate is $-0.93$ percentage points, with bootstrap interval $[-1.85,0.00]$; against Uniform-4 the interval is $[-4.17,1.39]$. Neither comparison establishes equal success probabilities.

Exploratory strata show that the spatial scan reduction against Anytime-gate is 14.15\% on DAO maps (both 108/108 arrivals) and 10.26\% on maze maps (98/108 versus 100/108 arrivals). Across increasing $\sigma=0.6,0.9,1.2$, the reductions are 1.25\%, 5.64\%, and 18.48\%, respectively, with 72 scenarios per noise level. Means here include failed episodes, so the noise pattern must not be interpreted as an isolated sensing effect; both additional spatial timeouts occur in mazes.

\begin{figure}[tbp]
\centering\includegraphics[width=\linewidth]{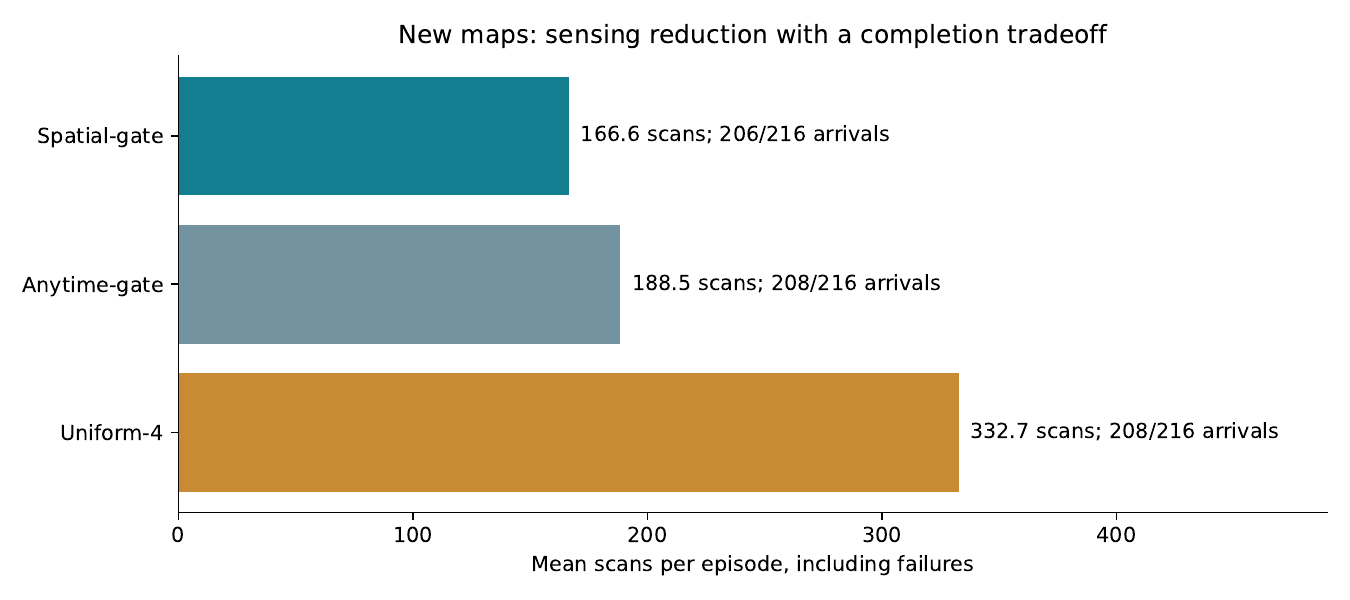}
\caption{Follow-up sensing costs with completion counts shown alongside them. The acquisition improvement accompanies two fewer arrivals relative to either principal reference.}
\end{figure}

\begin{figure}[tbp]
\centering\includegraphics[width=\linewidth]{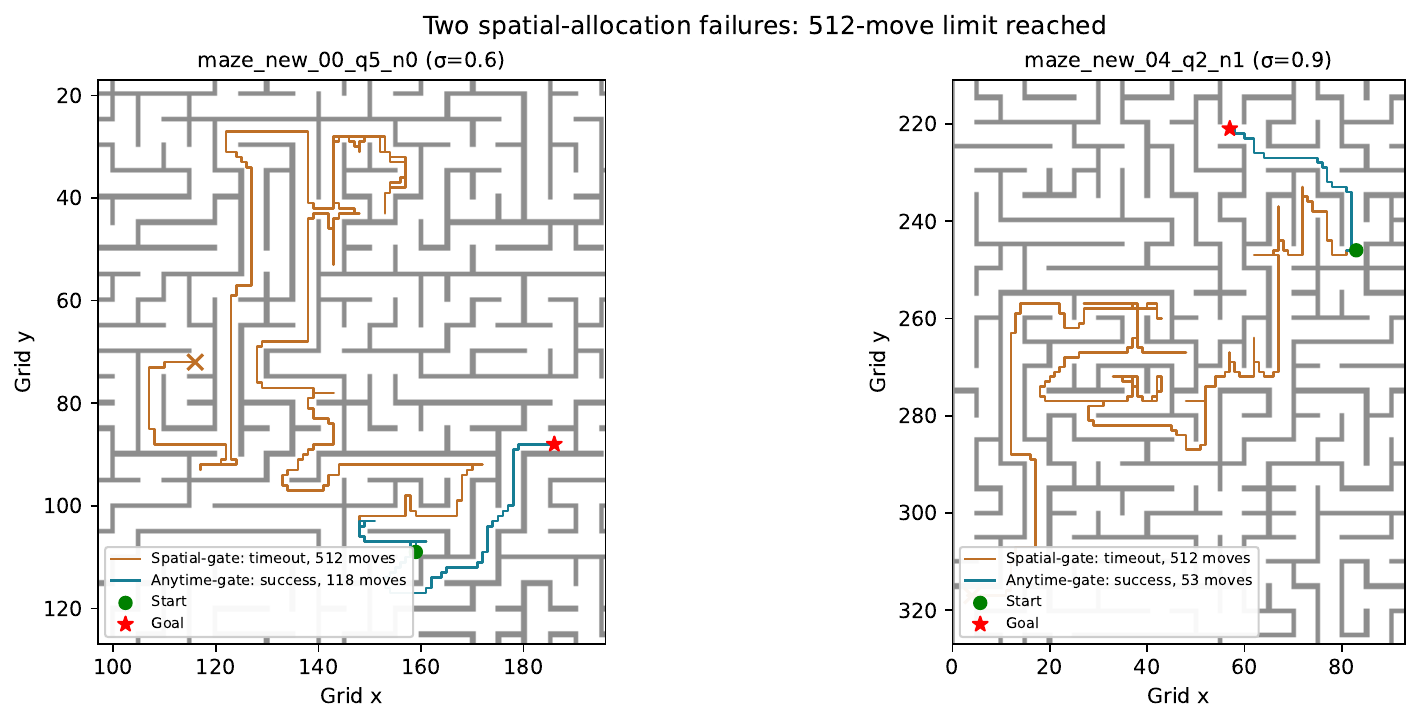}
\caption{Both additional Spatial-gate failures relative to Anytime-gate. The paths are deterministic replays on the new maze parents; orange trajectories stop at the 512-move limit. The blue paths reach the goal. These two cases are the complete set of discordant mission outcomes in this pairwise comparison.}
\label{fig:failures}
\end{figure}

The two missions that Spatial-gate fails and Anytime-gate completes both reach the 512-move cap. The first uses 512 moves and 594 scans against 118 moves and 130 scans for Anytime-gate; the second uses 512 moves and 958 scans against 53 moves and 109 scans. No mission shows the opposite success pattern. Figure~\ref{fig:failures} displays both complete trajectories. The allocation changes which cells become easier or harder to certify, but these traces alone do not isolate the ensuing routing and replanning decisions as a unique cause.

Deterministic replay shows that the paths first diverge after 20 and two common moves, respectively. Their scan histories have already diverged by those points (23 versus 26 scans; seven versus 12). At each mission's start and goal, the spatial weight is about 122 or 114 times its uniform counterpart, lowering the free-space evidence threshold from 15.47 to 10.67 or 10.74. These observations show that the allocation changes local certification and subsequent experience early, but do not identify a single causal step behind the later detours.

Among 206 joint successes with Anytime-gate, the scan reduction is 18.14\% (95\% interval: 15.70--21.25\%). Among 204 joint successes with Uniform-4, it is 55.68\% (51.91--58.50\%). These conditional analyses show that early termination alone does not explain the acquisition advantage, but do not remove the completion penalty.

Mean measured computation including spatial-allocation setup is 39.64\,ms per episode for Spatial-gate, versus 27.10\,ms for Anytime-gate and 30.20\,ms for Uniform-4. Setup alone averages 8.39\,ms. These measurements describe computation on the test host. The spatial variant uses more CPU time despite taking fewer scans.
\subsection{Robustness stress tests}
The original study also evaluated two deliberately difficult sensor settings on 54 scenarios each, drawn from the initial test parents. These are distinct from the 324-scenario nominal benchmark and were not used to select the principal configurations. The stronger independent-noise condition uses $\sigma=1.5$ with the corresponding known variance in the update. The correlated condition uses $\sigma=1.2$ and persistent per-cell bias at $\rho=0.5$, which violates the conditional-independence premise of the certificate. Table~\ref{tab:stress} counts every terminal outcome; the average scans include incomplete episodes.

\begin{table}[tbp]
\caption{Robustness stress tests on 54 episodes per condition and policy. Stop denotes the stationary verification limit; plan denotes planning failure. The correlated setting falls outside the nominal collision-bound assumptions.}
\label{tab:stress}\centering\small
\begin{tabular}{@{}llrrrrrr@{}}
\toprule
Condition & Policy & Arrive & Collide & Stop & Plan & Timeout & Scans\\
\midrule
Independent $\sigma=1.5$ & Anytime-gate & 40 & 0 & 14 & 0 & 0 & 162.11\\
 & Uniform-4 & 54 & 0 & 0 & 0 & 0 & 136.59\\
Persistent bias $\rho=0.5$ & Anytime-gate & 0 & 0 & 52 & 2 & 0 & 77.24\\
 & Uniform-4 & 30 & 7 & 0 & 14 & 3 & 477.22\\
\bottomrule
\end{tabular}
\end{table}

With stronger independent noise, the gate has no observed collisions but fails to complete 14 missions; a conditional collision bound does not guarantee progress. With persistent bias, its 52 verification stops and two planning failures preclude any completion claim, and the sequential certificate no longer has the stated validity. The fixed-scan reference completes 30 missions and collides in seven in that same biased setting. Spatial-gate was not subjected to these stress suites. These diagnostics do not establish robustness outside the nominal observation model.

\section{Ablation and Sensitivity Analysis}
\subsection{Certification error budget}
We reran the original 108 validation scenarios with the initial gate and the frozen routing weight $\lambda=1$, changing only the nominal episodic error budget $\delta$. This is a retrospective sensitivity diagnostic, not a new held-out test or a retuning of the reported test policies. The $\delta=0.05$ rerun reproduces the original validation episode outcomes and scan counts. Table~\ref{tab:delta} reports all outcomes, so a reduced scan mean cannot conceal a verification stop.

\begin{table}[tbp]
\caption{Retrospective error-budget sensitivity on 108 original validation scenarios. Routing weight, maps, noise seeds, and other execution settings are held fixed; scan means include all terminal outcomes.}
\label{tab:delta}\centering\small
\begin{tabular}{@{}rrrrrr@{}}
\toprule
$\delta$ & Arrivals & Collisions & Verification stops & Other failures & Mean scans\\
\midrule
0.01 & 108 & 0 & 0 & 0 & 144.15\\
0.02 & 107 & 0 & 1 & 0 & 142.50\\
0.05 & 108 & 0 & 0 & 0 & 133.81\\
0.10 & 108 & 0 & 0 & 0 & 129.74\\
\bottomrule
\end{tabular}
\end{table}

A larger budget lowers certification thresholds and generally reduces scan use in this set. The single stop at $\delta=0.02$ shows that episode outcomes need not be monotone in $\delta$: a changed gate can alter the subsequent route, observations, and use of the stationary scan limit. The numerical value of $\delta$ is a model-conditional upper bound, not an estimated collision frequency; observing no collisions in 108 runs does not validate the bound's assumptions.

\subsection{Spatial-allocation width}
The three widths were compared on the same 108 validation scenarios while keeping $\delta=0.05$, the mixture coefficients of the spatial weights, and $\lambda=1$ fixed. All three completed 108 missions without an observed collision. Mean scans were 109.48 at $h=8$, 112.95 at $h=16$, and 116.92 at $h=32$ cells. The smallest width was chosen before evaluation on the new map parents. The subsequent held-out comparison in Section~6.2 measures the performance of that selected width; it does not provide a held-out ranking of widths.

\section{Discussion and Limitations}
Anytime-gate at $\lambda=1$ used about 101 fewer scans per episode than Uniform-4 at $\lambda=4$, with both completing all 324 missions. Because their routing weights differed, some of the difference could reflect route selection. Matching the routing weights left a reduction of roughly 43--44\% at both tested weights (Table~\ref{tab:matched}). These comparisons hold the routing weight fixed within each pair. They were designed after the original results were known and use the same test scenarios.

The spatial variant redistributes the same total error allowance using the start and goal of each mission. Its weights favor cells near the initial start--goal rectangle and are fixed before any observation. The mechanism lowers the evidence threshold in favored regions, but the new-map test also produces two additional timeouts and higher CPU cost. A lower scan count is therefore only one part of its outcome.

The simulator counts scans and records component CPU time. The effect of fewer scans on energy use and elapsed mission time depends on sensor power and communication delays, which the simulation does not model. The 45.05\% reduction in the initial test and the 49.91\% reduction in the follow-up come from different map sets. The direct estimate for spatial allocation is the 11.58\% scan reduction relative to Anytime-gate on the same follow-up scenarios.

\subsection{Scope of the guarantee and next steps}
The guarantee depends on a known conditional observation model, static occupancy, exact pose, correct footprint geometry, and an initially free footprint. Persistent temporal bias violates the evidence-process premise. The stress tests in Section~6.3 expose a completion penalty under stronger independent noise and a collapse in completion under persistent bias. Spatial-gate was not stress-tested, so no robustness improvement is claimed for it.

The initial and follow-up tests cover 18 and 12 independent maps, respectively. Eligible queries have reachable endpoints that accommodate the robot's footprint and a moderate shortest-path length. The follow-up covers maze and DAO maps after the room maps failed to meet the preparation quotas. Testing more independent maps and longer missions would help establish how far these findings transfer.

All baseline settings in the follow-up are inherited from the earlier validation set. This tests transfer of fixed configurations, not each method's best possible adaptation to the new domains. The reference set directly controls repeated sensing and verification, but does not implement the safe-exploration framework of Stephens et al. or the chance-constrained trajectory optimizer of Dawson et al. Consequently, the study supports a within-simulator comparison of execution gates, not a state-of-the-art ranking across navigation architectures.

The model excludes velocity, braking, localization drift, dynamic obstacles, and a learned physical sensor model. Future work should investigate bias-aware evidence and completion failures on separate development data, then evaluate frozen changes on fresh test parents. Allocation implementation could also be optimized, with timing verified independently.
\section{Research Ethics and Reproducibility}
All evaluations use public grid benchmarks and synthetic measurements, without physical-robot trials. Benchmark attribution and applicable reuse terms should be preserved. The MovingAI collection recognizes BioWare's permission to distribute its maps for research \cite{sturtevant}. Simulation findings should not be used as a physical deployment certificate.

Implementation uses Python, NumPy, SciPy, pandas, Numba, and scikit-learn. The accompanying package records source archives and hashes, complete map and scenario manifests, random seeds, calibration parameters, validation candidates, frozen configurations, and all episode outcomes. The initial study contains 6,048 navigation episodes including tuning and stress suites; the follow-up adds 2,268 including width selection. The retrospective error-budget diagnostic adds 432 validation reruns. These totals include repeated runs on the same maps.

Fourteen initial-study checks and five follow-up tests passed, covering planning, sensing, geometry, certificates, map integrity, and frozen inputs. Executed notebooks capture actual outputs and deterministic replays. Their default execution validates stored experiments; separate commands rerun all episodes.

\section{Conclusion}
Sequential footprint verification reduced average acquisitions from 224.13 to 123.15 scans on the initial held-out test, while both principal policies completed 324/324 missions. On a separate set of maps, spatial allocation lowered the original gate's average scans from 188.45 to 166.63, but finished 206 rather than 208 of 216 missions and increased measured computation.

Under the nominal sensor model, sequential footprint verification can substantially reduce the number of scans while maintaining observed completion on the initial test set. The follow-up shows why both outcomes matter: spatial allocation saved additional scans, but two missions that the original gate completed reached the move limit. A next step is to develop evidence rules that account for persistent bias and test them on fresh maps, with scan counts and completion reported together.

\section*{Data and Code Availability}
Source maps and queries are available through MovingAI \cite{sturtevant}. The accompanying research archive contains the code, provenance records, raw outcomes, environment specifications, and executed notebooks. A public, citable location for the archive is pending.

\section*{Acknowledgments}
The authors acknowledge the public benchmark resources provided by the MovingAI Lab and the original map contributors. The work was prepared in the academic context of Euromed University of Fez.

\end{document}